%% file: main.tex
\documentclass[10pt,twocolumn,letterpaper]{article}

\usepackage[pagenumbers]{cvpr} % To force page numbers, e.g. for an arXiv version

\input{preamble}

\definecolor{cvprblue}{rgb}{0.21,0.49,0.74}
\usepackage[pagebackref,breaklinks,colorlinks,citecolor=cvprblue]{hyperref}

\usepackage{graphicx}
\usepackage{amsmath}
\usepackage{amssymb}
\usepackage{booktabs}
\usepackage{microtype}
\usepackage{adjustbox}
\usepackage{listings}
\usepackage{pythonhighlight}
\usepackage{float}
\usepackage{tablefootnote}
\usepackage{tabularx}
\usepackage{wrapfig}
\usepackage{duckuments}
\usepackage[T1]{fontenc}
\usepackage{wrapfig}
\usepackage{amssymb}

\usepackage[normalem]{ulem}

\newcommand{\1}[1]{{\color{green}#1}}
\newcommand{\2}[1]{{\color{blue}#1}}
\newcommand{\ig}[1]{{\color{red}#1}} % ignore

\def\paperID{7693} % *** Enter the Paper ID here
\def\confName{CVPR}
\def\confYear{2024}

\title{FaceLift: Semi-supervised 3D Facial Landmark Localization}
\author{David Ferman  
\and
Pablo Garrido  
\and
Gaurav Bharaj\\
\and
\begin{tabular}{ccc}
~~~~~~~~~~~~~~~~~~~~ & Flawless AI & ~~~~~~~~~~~~~~~~~~~~
\end{tabular}
}

\begin{document}
\maketitle
\input{sec/0_abstract}    
\input{sec/1_intro}
\input{sec/2_relatedwork}
\input{sec/3_method}
\input{sec/4_experiments}
\input{sec/5_conclusion}

% {
%     \small
%     \bibliographystyle{ieeenat_fullname}
%     \bibliography{main}
% }

{
    \small
    \bibliographystyle{ieeenat_fullname}
    \bibliography{main}
}
\clearpage
\appendix
\input{sec/6_supplementary}

% \clearpage
% {
%     \small
%     \bibliographystyle{ieeenat_fullname}
%     \bibliography{main}
% }

\end{document}

%% file: preamble.tex
\usepackage[dvipsnames]{xcolor}
\newcommand{\red}[1]{{\color{red}#1}}

%% file: sec/0_abstract.tex
\begin{abstract}
% Sentence 1: CONTEXT - why now?
3D facial landmark localization has proven to be of particular use for applications, such as face tracking, 3D face modeling, and image-based 3D face reconstruction.
% Sentence 2: NEED - why does the reader care?
In the supervised learning case, such methods usually rely on 3D landmark datasets derived from 3DMM-based registration that often lack spatial definition alignment, as compared with that chosen by hand-labeled human consensus, \eg,~how are eyebrow landmarks defined? This creates a gap between landmark datasets generated via high-quality 2D human labels and 3DMMs, and it ultimately limits their effectiveness.
% % Sentence 3: TASK - what do we do?
% % Sentence 4: OBJECT - what does this document do?
To address this issue, we introduce a novel semi-supervised learning approach that learns 3D landmarks by directly lifting (visible) hand-labeled 2D landmarks and ensures better definition alignment, without the need for 3D landmark datasets. To lift 2D landmarks to 3D, we leverage 3D-aware GANs for better multi-view consistency learning and \emph{in-the-wild} multi-frame videos for robust cross-generalization.
% \red{\sout{Furthermore, we contribute a novel 3D facial landmark evaluation scheme to handle comparison across various 3D landmark definitions by exploiting recent advancements in photogrammetric face mesh tracking.}}
% % Sentence 5: FINDINGS - what did we discover?
% % Sentence 6: CONCLUSIONS - so what?
% % Sentence 7: PERSPECTIVES - what now?
Empirical experiments demonstrate that our method not only achieves better definition alignment between 2D-3D landmarks but also outperforms other supervised learning 3D landmark localization methods on both 3DMM labeled and photogrammetric ground truth evaluation datasets.
Project Page: \href{https://davidcferman.github.io/FaceLift}{https://davidcferman.github.io/FaceLift}
\end{abstract}

%% file: sec/1_intro.tex
\section{Introduction}
\label{sec:intro}

\begin{figure}
    \centering
    \includegraphics[width=1.0\columnwidth]{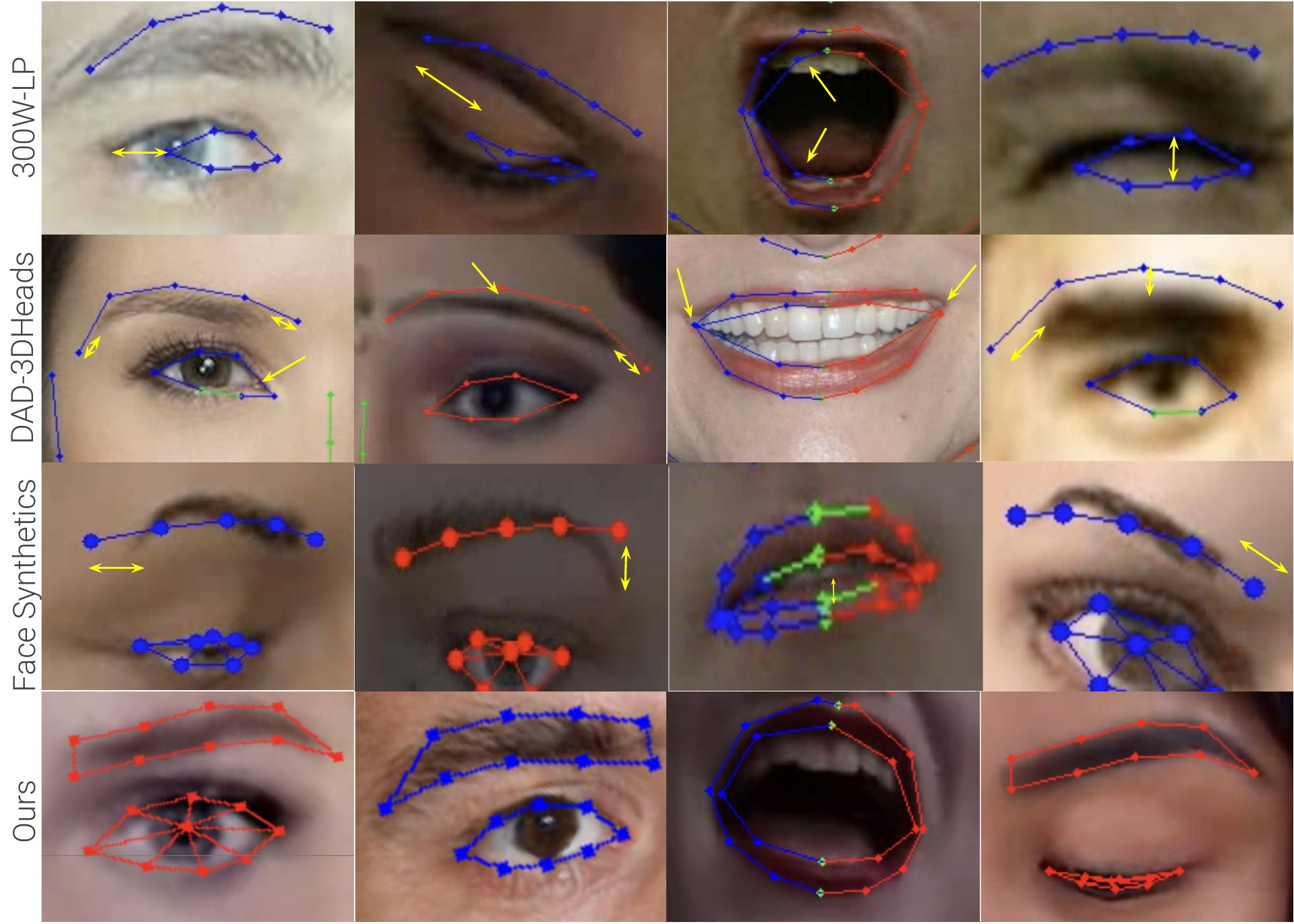}
    \caption{Comparison of our labels with 300W-LP~\cite{zhu2016_face-align}, DAD3D-Heads~\cite{martyniuk2022_dad-3dheads}, which are both labeled via 3DMM, and Microsoft's Face Synthetics~\cite{wood2021_fakeit} datasets.}
    \label{fig:3DMM}
\end{figure}

3D facial landmark localization plays a critical role in various applications, such as talking head generation~\cite{wang2021_one-shot}, 3D face reconstruction~\cite{martyniuk2022_dad-3dheads,guo2020_towards-fast,zhu2019_face-align}, and learning 3D face models~\cite{zheng2022_imface}. However, existing 3D facial landmark datasets based on 3D Morphable Model (3DMM) often lack alignment with 2D landmark definitions labeled by humans. This leads to a noticeable ambiguity between 2D and 3D datasets and limits their overall effectiveness, as shown in~\cref{fig:3DMM}. We propose an algorithm to bridge this ambiguity by directly lifting hand-labeled 2D landmarks into 3D, without additional 3D landmark localization datasets.

Human-labeled 2D datasets are known to exhibit high-quality facial landmarks for visible facial regions, while self-occluded regions are labeled in a ``landmark-marched'' style~\cite{zhu2015_high-fidelity}, i.e., on the nearest visible boundary. On the other hand, current 3D datasets leave much to be desired in terms of accuracy and consistency w.r.t 2D landmark definitions. For example, human-labeled 2D facial landmark datasets focus on the apparent brow boundaries, whereas 3DMM-based models define the brow region structurally above the eyes, as fixed mesh vertices. However, the relationship between facial structure and brow appearance varies across identities, and hence, a 2D-3D inconsistency occurs, see~\cref{fig:3DMM}.

\begin{figure*}[!htb]
    \centering
    \includegraphics[width=1.0\textwidth]{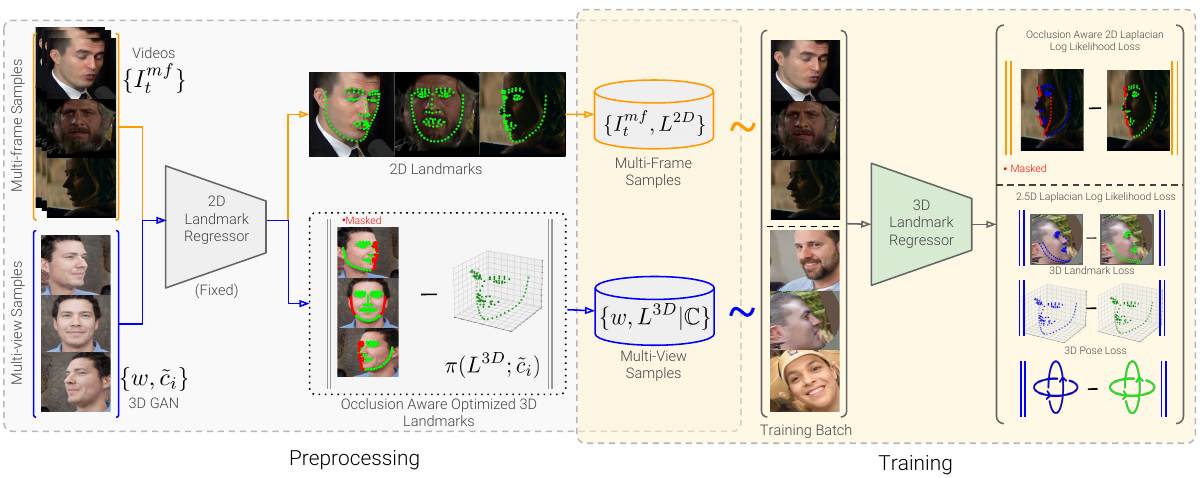}
    \caption{\textbf{System Pipeline:} We preprocess multi-frame videos, $\{I_t^{mf}\}_{t=1}^T$, and multi-view 3D-aware GAN samples, $\{I^{mf}_i=GAN(w; \tilde{c_i})\}_{i=1}^{|\tilde{C}|}$, predicting 2D landmarks for each image. For each GAN latent $w$, we optimize a set of 3D landmarks to minimize a masked, occlusion-aware, reprojection error across views $\tilde{c}_i \in \tilde{C}$, to obtain 3D pseudo-labels. Next, we train a 3D landmark regressor on batches of 2D pseudo-labeled multi-frame samples and 3D pseudo-labeled multi-view 3D-aware GAN samples, supervising via a combination of 2D confidence-aware losses and 3D landmark and pose losses, masking the 2D pseudo-labels in an occlusion-aware manner.}
    \label{fig:pipeline}
\end{figure*}
Inconsistencies are particularly evident in fine-scale details not captured by the linear 3DMM fitting, often seen in the mouth and eyes where fine-scale details are crucial for accurate representation, as noted by ~\cite{martyniuk2022_dad-3dheads}, see~\cref{fig:3DMM}. Additionally, unlike 2D methods, the SoTA models trained on such datasets tend to fail to capture blinks, as shown in~\cref{fig:result}. Finally, ``hallucinated'' self-occluded landmarks are prone to labeling errors due to the difficulty in labeling the non-visible regions~\cite{zeng2023_3d-aware}. We observe that the ``visible'' subset of 2D landmarks is fairly 3D consistent, see~\cref{fig:maskedMV}, i.e., they mimic what we refer to as ``2.5D'' (projected from 3D) landmarks. Inspired by these observations, we investigate whether it is feasible to lift visible 2D landmarks into 3D.

Thanks to recent advancements, volumetric 3D-aware GANs have enabled the generation of synthetic, yet photorealistic, multi-view images with controllable ground-truth camera information. Despite remarkable progress, we observe that available methods are still imperfect w.r.t. multi-view appearance consistency~\cite{chan2022_eg3d, sun2022_ide}, while improved modeling is on-going~\cite{an2023_panohead}. In view of the present limitations, we hypothesize that we can exploit existing 3D-aware GANs as a 3D prior and 2D landmarks as 2D image constraints to reveal the 3D awareness of human faces while preserving the 2D-3D consistency. %\gb{To the best of our knowledge, we are the first to achieve this 2D-3D consistency and thus enable 3D landmark localization consistent with 2D human-defined labels.}

In order to obtain 2D-3D consistent 3D landmarks, we propose a semi-supervised approach for 3D landmark detection, which leverages 1) a 3D-aware GAN prior for multi-view and multi-frame information from \emph{in-the-wild} videos and 2) 2D landmark pseudo-labels\footnote{Non-ground-truth labels} from a SoTA 2D detector. Our method trains jointly on multi-frame samples, pseudo-labeled by the 2D landmark detector, and on multi-view samples, with 3D pseudo-labels obtained via lifting 2D detections from multiple views. Training purely on multi-view 3D-aware GAN samples would introduce a bias and lack of sufficient variation in lighting, image quality, and facial expressions due to the limited diversity of the dataset, FFHQ~\cite{karras2019_style} and its extrapolation, while \emph{in-the-wild} videos contain such diversity. 

As previously noted, 3D-aware GAN sampled data, in its current state, is still imperfect, as we observe certain fine-scale details can vary with pose, especially for features like eyelids and pupils, while large poses often lead to severe appearance degradation and background boundary artifacts. While multi-frame samples from videos contain rich diversity, we cannot rely on these samples exclusively as they lack the 3D constraints offered by the multi-view 3D-aware GAN samples, and only a subset of 2D landmarks are 2D-3D consistent, as previously noted. Additionally, while \emph{in-the-wild} videos are biased toward frontal camera-facing head poses~\cite{zhu2022_celebv}, sampling from a 3D-aware GAN offers full controllability over the 3D pose distribution, offering more balanced training. Thus, by combining the merits of multi-view and multi-frame samples we are, to the best of our knowledge, the first to achieve this 2D-3D consistency and thus enable 3D landmark localization consistent with 2D human-defined labels.

% While aiming for a qualitatively better 3D landmark definition, we ensure landmark accuracy and 3D consistency w.r.t. ground-truth are not compromised, necessitating a methodology for comparing our method, that differs in landmark definition with existing approaches. With the availability of high-quality ground-truth temporally consistent 3D mesh tracking datasets such as Multiface~\cite{wuu2022_multiface}, we introduce an evaluation scheme for 3D landmarks that is agnostic to the landmark definition, enabling such a comparison. For performance evaluation on \emph{in-the-wild} data, we utilize the same scheme on the recently introduced DAD-3DHeads dataset.
We evaluate our method on the \emph{in-the-wild} DAD-3DHeads dataset and on high-quality ground-truth temporally consistent 3D mesh tracking dataset, Multiface~\cite{wuu2022_multiface}.
%we also employ our evaluation scheme using the recently introduced DAD3D-Heads dataset.
On both datasets~\cite{wuu2022_multiface,martyniuk2022_dad-3dheads}, we achieve state-of-the-art accuracy when comparing to existing SoTA methods, despite being trained without a ground-truth 3D dataset. %\gb{In this paper, we propose a novel semi-supervised approach for high-quality 3D landmark localization that leverages the multi-view consistency of high-quality ``visible'' 2D landmarks.} 
To summarize, our main contributions are as follows:
\begin{enumerate}
    \item We introduce a semi-supervised approach that leverages high-quality 2D landmarks along with a 3D-aware GAN prior to tackle the 2D to 3D lifting problem. The resulting pipeline is geometric prior free, enabling learning accurate 3D landmarks that align with 2D hand-labeled definitions, without any ground-truth 3D labels.
    \item A novel 3D transformer formulation that leverages volumetric consistency (multi-view constraint) while training on real videos (multi-frame constraint) for \emph{in-the-wild} generalization.
    % why we use the multi-view and multi-frames?
    % \item \red{\sout{A novel landmark definition agnostic evaluation scheme comparing different 3D landmark detectors via optimal mesh vertex subset selection for each detector.}}
    \item State-of-the-art accuracy on photogrammetric ground-truth Multiface~\cite{wuu2022_multiface} and human-labeled DAD-3DHeads~\cite{martyniuk2022_dad-3dheads} datasets, achieving cross-generalization.
\end{enumerate}

%% file: sec/2_relatedwork.tex
\section{Related Work}
\label{sec:relatedwork}
\begin{figure*}
    \centering
    \includegraphics[width=1.0\textwidth]{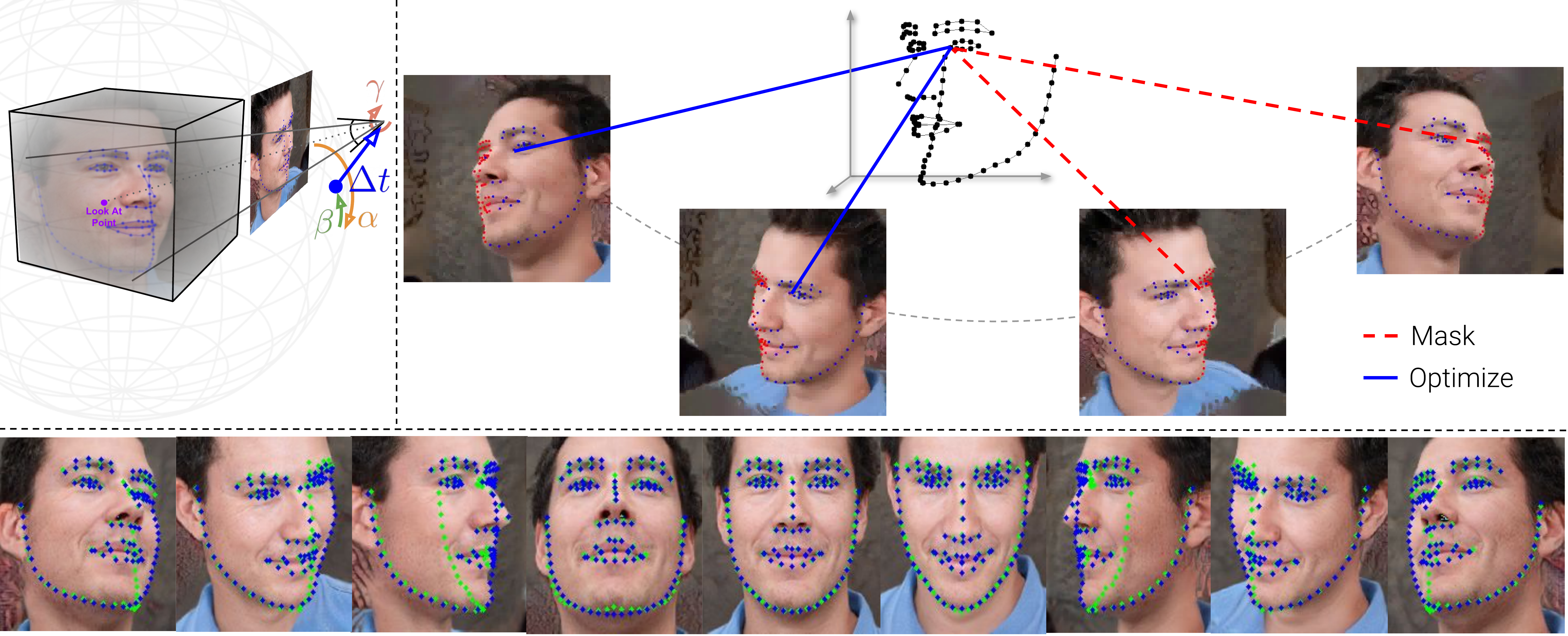}
    \caption{\textbf{Masked Multiview 3D Landmark Optimization:} Top Right: We define a fixed set of camera views and hand-design view-dependent landmark masks based on 3D landmark visibility and 2D landmark detector competency. For a given identity sampled from the 3D GAN, we render the set of views, predict landmarks via the 2D detector, and optimize 3D landmarks to minimize the view-dependent-masked reprojection error across all views. Top Left: Rendering of 3D-aware GAN with camera $c$ parameterized by azimuth, elevation, and roll angles $\alpha, \beta, \gamma$ along a sphere, and translation $\Delta \mathsf{t}$. Bottom: Illustration of 2D-3D consistency between optimized 2.5D projections of 3D pseudo-labels, green, and 2D landmark detections, blue.}
    \label{fig:maskedMV}
\end{figure*}

We review methods for 2D-to-3D pose and keypoint estimation and 3D facial landmark localization. In addition, we discuss existing facial landmark datasets.
% \begin{itemize}
%     \item  3D Human Body Pose Uplifting: Human Pose Estimation (HPE), w.r.t. uplifting there is a line of work, so we have to select the representative works.} first predicts 2D keypoints and then lifts those keypoints, whereas we lift in the volumetric space during preprocessing rather than forward pass.
%     \item Bulat Uplifting
%     \item Note: All human body keypoints already exist in what we refer to as 2.5D, while 2D facial landmark datasets are not labeled with 2D-3D correspondences.
% \end{itemize}
\paragraph{2D-to-3D Uplifting for Pose and Landmark Estimation.} Direct estimation of 3D pose and landmarks from images is an ill-posed problem~\cite{li2014_3dhuman}, and ground truth 3D image annotations are often limited~\cite{pavlakos2018_ordinal}. As such, methods in this category often require 3D priors~\cite{li2014_3dhuman, guo2020_towards-fast, wood2022_3dface} or depth supervision~\cite{pavlakos2018_ordinal}. On the other hand, lifting methods leverage intermediate representations, such as 2D pose or landmark detectors~\cite{martinez2017_simple-effective,bulat2017_2d-3d,zheng2021_3d-human-pose}, or temporal information e.g. via graph convolutional networks~\cite{liu2020_comprehensive,zhao2019_semantic,ma2021_context} to infer 3D information. Due to the excellent performance of 2D detectors~\cite{cao2021_openpose,ferman2022_mdmd,sun2019_deep-hrnet}, 2D-to-3D uplifting methods normally outperform direct 3D regression methods. Interestingly, while 2D-to-3D pose uplifting has been investigated more extensively, almost no work for face landmark estimation has been done~\cite{bulat2017_2d-3d,zhang2019_adversarial}, mainly due to the wide availability of 3D face priors~\cite{egger2020_3d-morphable} and recent photo-realistic synthetic datasets~\cite{zhu2016_face-align,wood2021_fakeit}. We note that the definitional gap between 2D and ``2.5D'', see bottom~\cref{fig:maskedMV}, presents an additional challenge for uplifting facial landmarks, as 2D labels cannot be modeled simply as projections from 3D, as in the case of human pose estimation.
Despite their impressive performance, 2D-to-3D uplifting remains an inherently ill-posed problem, even when spatio-temporal modeling is adopted, since multiple solutions are available, especially when occlusions occur~\cite{li2022_mhformer}. Recently, transformer-based methods have been introduced, which exploit attention to better reason over temporally neighboring 2D poses for temporal-aware lifting ~\cite{zheng2021_3d-human-pose, li2022_mhformer}. While these methods 
%utilize transformer decoders to 
attend to relevant temporal information for lifting to 3D, the problem of 3D facial localization lacks temporal ground-truth datasets and remains unexplored. To the best of our knowledge, no work for 2D-to-3D face landmark uplifting has been explored with 3D transformer architectures without ground truth 3D datasets.

\paragraph{3D Facial Landmark Detection on Images.} Methods for 3D landmark estimation can be categorized as template-based, 3DMM-based, 3D aware, and 2D-to-3D uplifting.
Template-based approaches exploit the template's underlying mesh topology for predicting spatial deformation maps in UV texture space~\cite{feng2018_joint-3d, piao2019_semi, ruan2021_sadrnet} or dense 3D face deformations~\cite{bhagavatula2017_faster-align, tewari2019_fml}.
3DMM-based methods utilize a 3D face model, e.g., BFM~\cite{paysan2009_3d-face} or FLAME~\cite{li2017_flame}, directly to estimate model parameters~\cite{wood2022_3dface}, often with surrogate 2D landmark supervision~\cite{guo2020_towards-fast,martyniuk2022_dad-3dheads}, or as an intermediate representation for 3D landmark refinement~\cite{zhu2016_face-align,zhu2019_face-align,wu2021_synergynet}.
%Landmark refinement approaches adopt cascaded regression with PNCC-based convolutions\cite{zhu2016_face-align} and additionally pose adaptive convolutions\cite{zhu2019_face-align}, multiple feature aggregation, e.g., shape, expression, and image features, in a cyclic prediction framework \cite{wu2021_synergynet}, or fusion of multi-scale neural features and landmarks.
%\cite{zhu2016_face-align} convolve image features with projected normalized coordinate code (PNCC) to refine model parameters. \cite{zhu2016_face-align} extend it with a two-stream refinement that adopts both pose adaptive convolutions and PNCC-based convolutions.
%\cite{guo2020_towards-fast} introduce a surrogate 2D landmark prediction task to regularize landmark regression and a video synthesis augmentation to improve tracking stability in videos.
%Similarly, \cite{wu2021_synergynet} aggregate shape, expression, and image features to refine model-based landmark regression.
%\cite{martyniuk2022_dad-3dheads} fuse coarse heatmap-based landmark prediction and multi-scale neural feature to jointly optimize pose and 3D landmark locations.
%
While template- and model-based methods have demonstrated robustness for 3D landmark localization, the representation power is limited by the underlying 3D dense prior~\cite{tewari2019_fml,mallikarjun2021_learning}.
3D aware techniques leverage volumetric representations to embed 3D landmarks~\cite{zhang2022_flnerf} or generate explicit multi-view image constraints for 3D consistent landmark prediction~\cite{zeng2023_3d-aware}. We note that both of these methods require 3D GAN inversion of monocular 2D images, either at inference or training, which is known to fail for large poses and occlusions~\cite{xie2022high}. Rather than inverting images, we lift 2D landmarks into 3D by exploiting the multi-view information of 3D-aware GAN samples, avoiding errors introduced by inversion.
Unlike previous approaches, 2D-to-3D uplifting methods require no geometry prior and directly regress 3D landmarks~\cite{bulat2017_2d-3d} or 3D shape consistent landmarks with moving boundaries via heatmap-based regression~\cite{zhu2015_high-fidelity,xiao2017_recurrent}. Alternatively, joint coordinate and adversarial voxel regression have been proposed~\cite{zhang2019_adversarial}.
%While heatmap-based methods exploit the visual similarities between facial components using local spatial and contextual information to reason about the location of landmarks~\cite{bulat2021_subpixel}, they often are less practical for modeling true 3D-like landmarks.
%Concurrently, state-of-the-art 2D heatmap-based methods also focus on predicting 2D "landmark-marched" contour lines ~\cite{zhu2015_high-fidelity,xiao2017_recurrent} that respect visible boundary lines while maintaining overall face shape consistency, or build upon boundary-focused approaches~\cite{wu2018_look,huang2021_adnet} that are designed around exploiting these visible boundaries.
As far as we are aware, the use of transformer-based 3D architectures without geometric priors for 3D sparse localization remains unexplored.

\paragraph{Facial Landmark Datasets.} A variety of 2D, 2.5D, and 3D face datasets have been proposed to advance research in facial landmark localization. 2D datasets contain ground truth 2D landmarks annotated on real images~\cite{liu2019_high-efficiency,wayne2018_look-boundary,sagonas2013_300-faces, kostinger2011_aflw}. Here, visible landmarks are aligned to face image features, while object-occluded landmarks are hallucinated and self-occluded landmarks are snapped to image boundaries, thus destroying overall 3D face likeness. Kumar et al.~\cite{kumar2020_luvli} partially solve this problem by labeling landmarks with three visibility categories: visible, externally-occluded, and self-occluded. However, these categories, and especially the latter, are created based on human perception, not metrics, and thus they are error prone.
2.5D datasets are either synthetically generated from rendered 3D meshes that attempt to bridge the photorealism gap~\cite{wood2021_fakeit} or derived from real images by automatically fitting a 3DMM~\cite{zhu2016_face-align}. 
Pure 3D datasets are derived from coarse 3DMM-based renderings~\cite{synthesisai2022_faces} or generated by densely fitting a 3DMM to real images with human supervision~\cite{martyniuk2022_dad-3dheads}.
Both 2.5D and 3D synthetic datasets have landmarks registered to specific 3D face mesh locations, which are not always aligned to 2D facial image features, e.g. eyebrows. As 2.5D and 3D real datasets are derived from 3DMM-based fitting, which is an ill-posed problem, perfect annotations cannot always be achieved, as shown in~\cref{fig:3DMM}.
While there exist smaller-scale multiview face datasets captured in controlled studio setups with photogrammetry data~\cite{cao2013_facewarehouse,wuu2022_multiface,bagdanov2011_florence,cosker2011_d3dfacs,yin2006_bu3d-fe,zhang2014_bp4d-s,zhang2016_bp4d-ms}, i.e., metrically accurate 3D reconstructions, these datasets are not applicable for in-the-wild facial landmark generalization.
Our semi-supervised approach overcomes limitations of 2D, 2.5D, and 3D datasets by leveraging the accuracy of visible 2D landmarks on real images and lifting them via 3D prior supervision with our novel 3D transformer formulation. Thus, our method requires no large-scale annotated 3D datasets, which to date are non-existent and nearly impossible to generate.

%% file: sec/3_method.tex
\section{Method}
We introduce a semi-supervised approach for learning 3D facial landmarks from a 3D-aware GAN prior and high-quality 2D landmarks~\cite{ferman2022_mdmd},
%\footnote{2D landmark detector is trained on SynthesisAI~\cite{synthesisai2022_faces}, WFLW~\cite{wayne2018_look-boundary}, and LaPa~\cite{liu2019_high-efficiency}.} \pg{Details go to the supp. material for completeness. Please explain why we use synthetic data.}
without the use of 3D labels, see~\cref{fig:pipeline}. Our method consists of a pre-processing stage and a training stage. We first pre-process our training data by predicting 2D landmarks on multiview 3D-aware GAN samples and \emph{in-the-wild} videos. The multiview landmarks from GAN samples are lifted to 3D via an occlusion-aware masked optimization to obtain 3D landmark pseudo-labels for each GAN latent. In our second phase, we train jointly on multi-view GAN samples, supervised by ground-truth 3D pseudo-labels, and multi-frame \emph{in-the-wild} videos, supervised via pose-dependently masked 2D pseudo-labels.

\paragraph{Pre-Processing}
We obtain data for training our method from multi-view 3D-aware GAN samples, along with multi-frame \emph{in-the-wild} videos. For each video frame, $I_t^{mf}$, we predict $N$ 2D landmarks, $L^{2D} \in \mathbb{R}^{N\times2}$, using a high-quality 2D landmark detector~\cite{ferman2022_mdmd}, which was trained on the WFLW~\cite{wayne2018_look-boundary} and LaPa~\cite{liu2019_high-efficiency} 2D landmark datasets, concurrently. For each GAN sampled latent code, $w$, we render a set of views and fit 3D landmark pseudo-labels via an occlusion-aware objective on multi-view 2D detections. In the following, we introduce the camera model of the 3D-aware GAN, and describe our landmark pseudo-label optimization and 3D landmark localization model.
%We explain the details of this optimization below, but first introduce the camera model of the 3D-aware GAN.\looseness=-1
\begin{figure*}
    \centering
    \includegraphics[width=\textwidth]{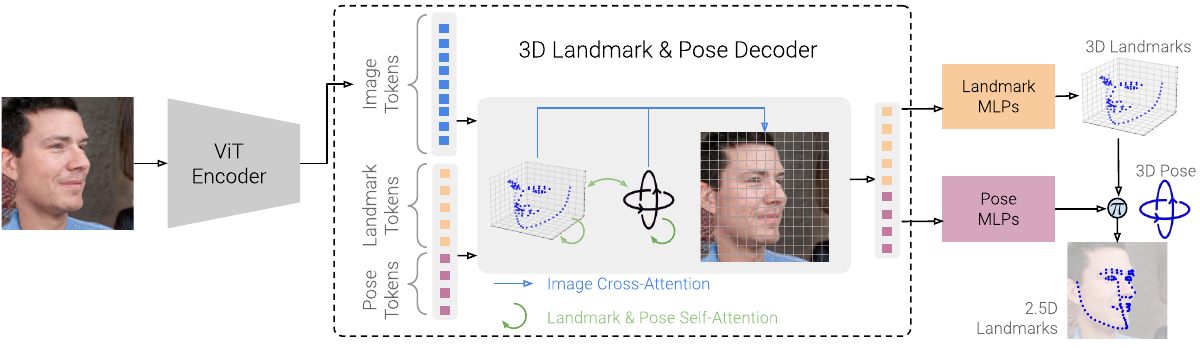}
    \caption{\textbf{3D Landmark Regressor Architecture:} Face images are embedded via a ViT encoder to obtain image tokens. Landmark and pose tokens are initialized from a learned embedding and passed through a 3D landmark and pose decoder, in which landmark and pose tokens cross-attend to the image tokens and perform self-attention over the ``sequence'' of landmark and pose tokens. Each landmark and pose token are routed to an MLP head to predict 3D landmarks and 3D pose, respectively. Finally, the 3D landmarks are projected to 2.5D landmarks via the predicted 3D pose.}
    \label{fig:3d}
\end{figure*}

\paragraph{Augmented Camera Space}
We share a perspective camera model~\cite{zollhofer2018_sota} between the volumetric rendering of the 3D-aware GAN and projecting 3D landmarks to screen space. Typically, volumetric face GANs use a camera with extrinsics  $\mathsf{M} = \left(\begin{smallmatrix} \mathsf{R} & \mathsf{t} \\ \ \mathbf{0} & 1 \end{smallmatrix}\right) \in \mathbb{R}^{4\times 4}$ parameterized by an azimuth angle, $\alpha$, and elevation angle, $\beta$, such that the camera is situated on a sphere pointing at the look-at point. We augment $\mathsf{M}$ with camera roll $\gamma$ and $\Delta \mathsf{t}$ applied to $\mathsf{R}$ and $\mathsf{t}$, respectively, see top-left~\cref{fig:maskedMV}. 3D-aware face GAN's also define camera intrinsics, $\mathsf{K} \in \mathbb{R}^{3\times3}$, with a fixed focal length, as described in~\cite{sun2022_ide}. Let $\mathbb{C}$ be the space of cameras projections s.t.\;$c=(\mathsf{K},\mathsf{M})\in \mathbb{C}$ iff $\alpha \in[-A,A], \beta \in[-B,B], \gamma \in [-\Gamma,\Gamma]$, and $\Delta \mathsf{t}$ such that the bounding box of projected facial landmarks is contained within the image and has minimum dimension greater than half the image dimension. % providing a sensible set of feasible combinations of $(R, t_{xy}, T_z)$.
A 3D landmark, $l^{3D} \in \mathbb{R}^3$, is projected from the GAN's canonical space to screen space via the perspective projection function $l^{2D}=\pi(l^{3D}; c)$:
\begin{align}
    \pi(l^{3D}; c)= [l^{2D}_x, l^{2D}_y, w]^\top / w; \\ [l_x^{2D}, l_y^{2D}, w]= \mathsf{K}\cdot(\mathsf{R} \cdot l^{3D} + \mathsf{t}).
\end{align}

\paragraph{Model Architecture}
We employ a transformer encoder-decoder model for predicting 3D landmarks, as shown in~\cref{fig:3d}. We use a ViT encoder~\cite{dosovitskiy2021_image-worth}, known as FaRL~\cite{zheng2022_general-rep}, pre-trained for human face perception tasks, which we show yields slightly better performance than Resnet152~\cite{he2016_deep-residual}. We design a transformer decoder with a token per landmark and pose tokens for rotation, Txy and Tz. These tokens pass through three blocks, each containing an image-cross-attention layer, landmark-pose self-attention layer, and MLP, with layer-normalization prior to each. Finally, we pass the landmark and pose tokens individually through MLP heads, which predict the 3D landmarks, Cholesky factorization of the 2D covariances of projected 2.5D landmarks, and the 3D rotation and translation. We apply the 3D landmark predictions as offsets to a template, defined as the landmark-wise mean of our 3D pseudo-labels obtained during pre-processing, to obtain 3D landmark predictions, $\hat{L}^{3D} \in \mathbb{R}^{N\times3}$. The pose is predicted via a 6D rotation representation, akin to~\cite{hempel2022_6d-rotation-representation} from which a rotation matrix, $\hat{\mathsf{R}}$, is extracted. From $\hat{\mathsf{R}}$, we compute the 3D translation to the camera sphere, $\bar{\mathsf{t}}_{\hat{\mathsf{R}}}$, and predict $\hat{\Delta \mathsf{t}}$ to obtain $\hat{\mathsf{t}}=\bar{\mathsf{t}}_{\hat{\mathsf{R}}}+\hat{\Delta{\mathsf{t}}}$. Finally, we form our predicted camera, $\hat{c}=[\mathsf{K},\hat{\mathsf{M}}]$, with fixed intrinsics, $\mathsf{K}$, and obtain 2.5D landmarks $\hat{L}^{2.5D}=\pi(\hat{L}^{3D}; \hat{c})$.

\paragraph{Training Methodology}
We train our 3D landmark detector jointly on multi-view GAN sampled images, and multi-frame \emph{in-the-wild} video frames. For multi-view image, $I^{mv}=GAN(w; c)$, we sample a latent code w from a set of pre-processed latents, along with a random camera $c\in C$. For each multi-frame sample, we sample a random video from our pre-processed video dataset followed by a random frame $I^{mf}$ from the video. Each batch of training consists of 4 multi-view samples, $(I^{mv},{L^{3D}}^*, c^*)$, and 4 multi-frame samples, $(I^{mf}_t, {L^{2D}}^*)$. We formulate our loss function as a combination of multi-frame and multi-view losses. The multi-view loss consists of a 2.5D uncertainty-aware landmark loss, 3D landmark loss, and 3D pose loss. We employ a Laplacian Log Likelihood (LLL) objective parametrized by predicted Cholesky factorization of landmark covariances, akin to~\cite{kumar2020_luvli,ferman2022_mdmd}. Such parameterization enables the energy landscape to adapt to noise caused by rendering artifacts and allows the model to weigh the loss for each landmark prediction based on its 2D anisotropic confidences. For 3D landmark loss, along with the translation loss, on $\Delta t$, we adopt mean-squared-error, while for 3D rotation, we follow head pose estimation work~\cite{hempel2022_6d-rotation-representation} and use geodesic loss. Thus, our multi-view loss is defined as:
\begin{equation}
    \mathcal{L}_{\text{mv}} = \mathcal{L}_{\text{MSE\_}L^{3D}} + \mathcal{L}_{\text{MSE\_}\Delta t} + \mathcal{L}_{\text{LLL\_}L^{2.5D}} + \mathcal{L}_{\text{Geo\_R}},
\end{equation}
Since the set of views encountered when training on \textit{in-the-wild} videos is not fixed, such as in the 3D pseudo-labeling optimization, we employ a simple heuristic for obtaining masks, $m\in \{0,1\}^{N}$. We define a template of normal vectors for each landmark, apply the estimated rotation to each normal, and threshold the dot product with the forward vector to obtain the mask. Thus, we supervise the multi-frame video samples via their 2D pseudo-labels, ${L^{2D}}^*$ as:
\begin{equation}
    \mathcal{L}_{\text{mf}} =  \sum_{i=1}^N m_n \cdot \mathcal{L}_{lll}({l^{2D}_n}^*, \hat{l}^{2.5D}_n; \Sigma_n),
\end{equation}
where $\Sigma_n$ refers to the covariance matrix obtained via predicted Cholesky factorization, and $L_{lll}$ denotes the 2D laplacian-log-likelihood. Refer to~\cite{kumar2020_luvli} for details. For video training, this anisotropic confidence weighted loss enables the energy landscape to adapt to systematic noise from the 2D detector, such as extreme pose samples where the 2D detector may fail, while GAN-based extreme pose samples are constrained via fixed 3D losses. Thus, our complete objective is: $\mathcal{L} = \mathcal{L}_{\text{mf}} + \mathcal{L}_{\text{mv}}$.
% \begin{equation}
%     \mathcal{L} = \mathcal{L}_{\text{mf}} + \mathcal{L}_{\text{mv}} 
%     \label{eq:multi_frame_loss}
% \end{equation}

%% file: sec/4_experiments.tex
\section{Results \& Analysis}

\paragraph{Training Implementation Details} We train our method jointly on \emph{in-the-wild} videos obtained from the CelebV-HQ~\cite{zhu2022_celebv} dataset, selecting the first 10K videos, along with GAN samples obtained from IDE-3D~\cite{sun2022_ide}, sampling latents from the first 10K random seeds. Our pre-processing stage takes roughly 3 days on a single A10G GPU with 24GB RAM to obtain pseudo-labels. Since geometric augmentations (e.g., scale and translation) would break our 3D ground truth under perspective projection, we render GAN-generated images on the fly during training, sampling cameras uniformly from augmented camera space, $\mathbb{C}$. To obtain a sensible pose distribution, we softly decrease extreme rotation angle combinations by accepting sampled rotations with probability $e^{-((\frac{\alpha }{A})^2+(\frac{\beta }{B})^2+(\frac{\gamma }{\Gamma })^2)}$, with $A=110, B=60, \Gamma=90$.  Our IDE-3D renders and video frame crops are 224x224, to match the input dimension of our FaRL~\cite{zheng2022_general-rep} backbone. We train with a learning rate of 1e-5 for 225 epochs, with the Adam~\cite{Kingma2015_adam} optimizer, decaying the learning rate exponentially by a factor of $0.9$ every 3 epochs, taking roughly 4 days on a single GPU machine. We overcome an IDE-3D artifact, where a large pose causes the background to occlude the face, by exploiting IDE-3D's semantic field to set the density of background points in the near half of the viewing frustum to $-\infty$, prior to rendering. We ignore GAN-rendered pupil landmarks during training, as we observe a bias where pupils tend to follow the camera, breaking multi-view consistency.
\begin{figure}
  \centering
  \includegraphics[width=1.0\columnwidth]{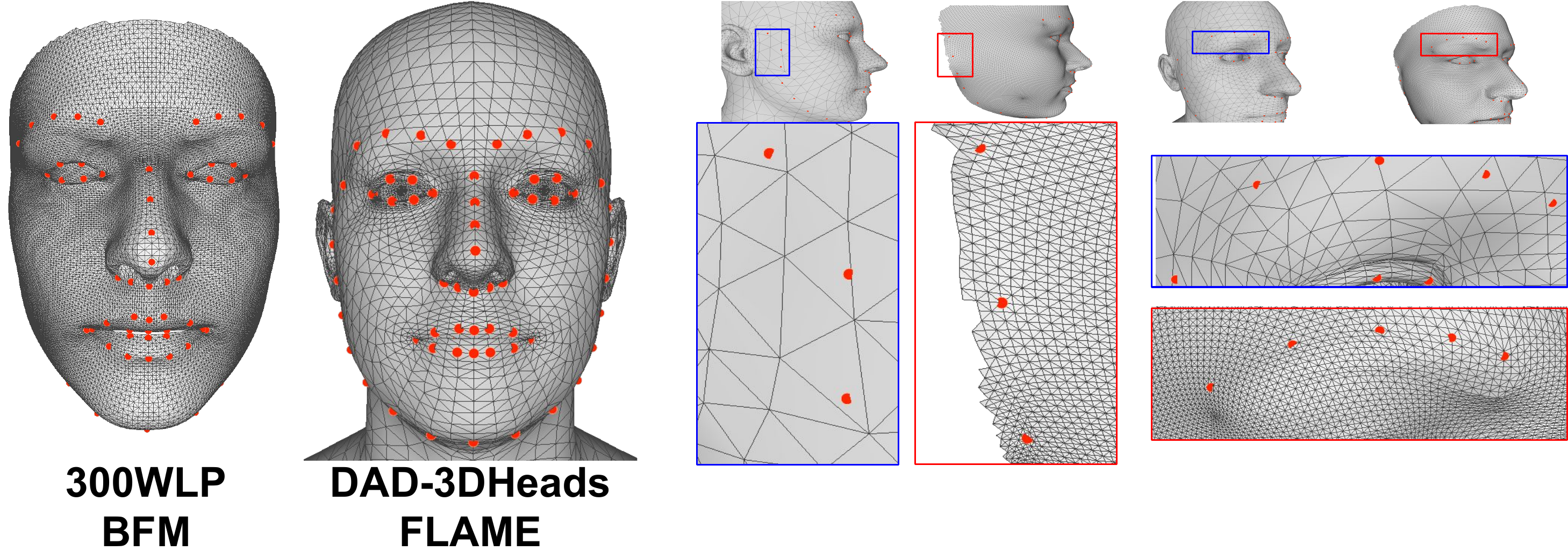}
  \caption{Global alignment of landmarks with local definition bias.}
  \label{fig:alignment}
\end{figure}
\begin{figure*}
    \centering
    \includegraphics[width=1.0\textwidth]{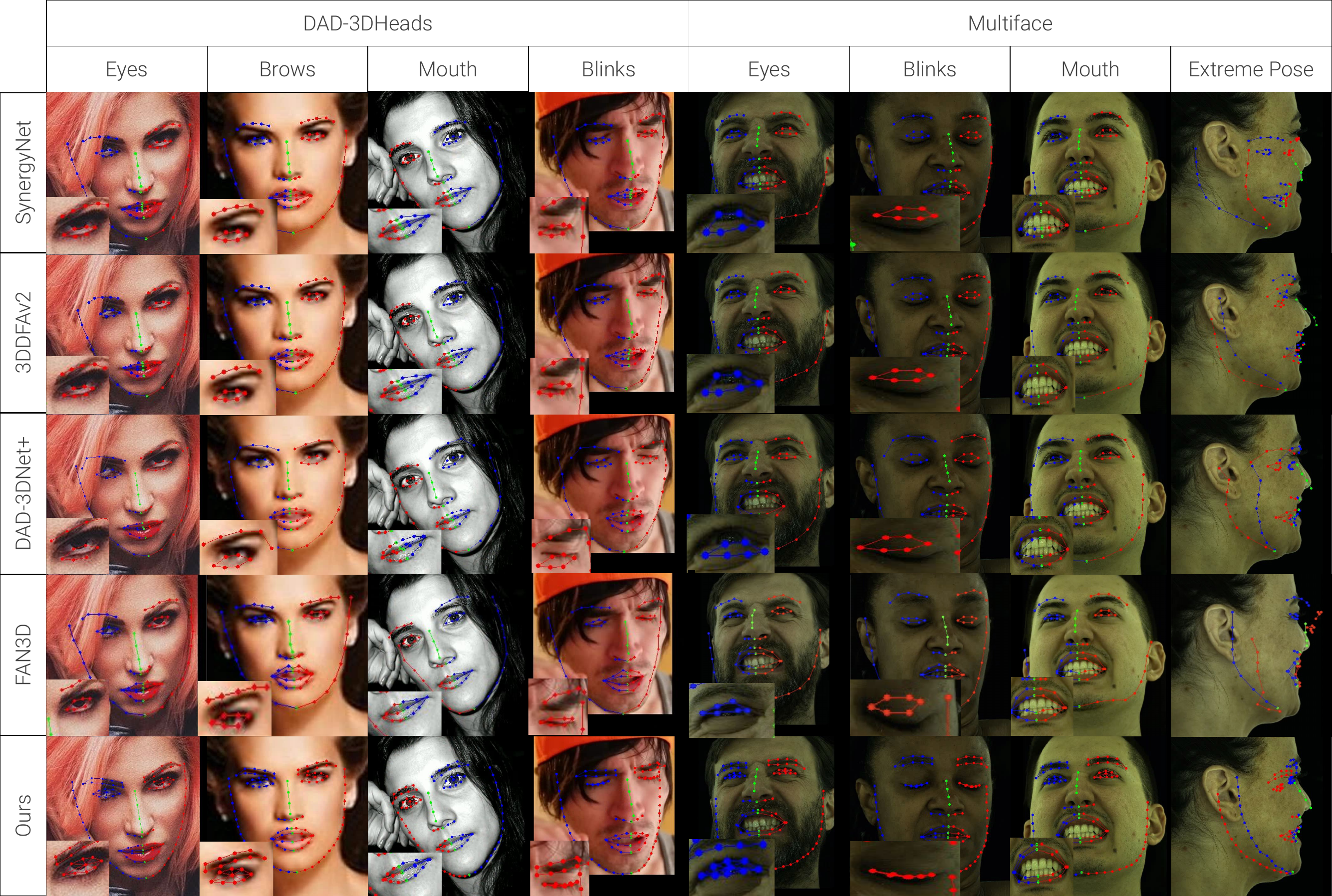}
    \caption{\textbf{Visual Results:} Ours, SynergyNet~\cite{wu2021_synergynet}, 3DDFAv2~\cite{guo2020_towards-fast}, DAD-3DNet+~\cite{zeng2023_3d-aware}, FAN3D~\cite{bulat2017_2d-3d} on DAD-3DHeads~\cite{martyniuk2022_dad-3dheads} and Multiface~\cite{wuu2022_multiface} samples.}
    \label{fig:result}
\end{figure*}

\paragraph{Normalized Mean Local Consistency Metric}
% 1. How landmarks are defined (vertices on face meshes of different top's or ours)
Previous works ~\cite{wu2021_synergynet,zhu2016_face-align,zeng2023_3d-aware,guo2020_towards-fast,bulat2017_2d-3d} train and evaluate $\text{NME}$ on datasets where selected indices of face mesh vertices define the landmarks. Across datasets, we observe that landmark definitions are globally aligned, i.e., same general semantic position, but suffer local definition bias, see~\cref{fig:alignment}, due to differences in vertex selections and mesh topology, e.g., 300WLP~\cite{zhu2016_face-align} uses BFM~\cite{paysan2009_3d-face} and DAD-3DHeads~\cite{martyniuk2022_dad-3dheads} uses FLAME~\cite{li2017_flame}, while ours does not use a mesh.
% 2. We can't compare locally, so we need a new metric (global alignment is assumed)
We report cross-dataset evaluations in the supplementary document, showing that traditional NME leads to unfair comparisons due to the local definition bias while still capturing a ballpark notion of global alignment. As such, we need a landmark definition agnostic metric for meaningful local comparison, which we introduce as an extension of standard $\text{NME}$.
% 3. NME used in previous works, can be thought of as parametrized by the indices that define landmarks (aka their definition)
We first define $\text{NME}$ as parametrized by the vertex indices. For a test set of $M$ images, with predicted landmarks $\hat{L}^{2.5D}_m \in \mathbb{R}^{N \times 2}$, projected vertex labels, $V^{2.5D}_m \in \mathbb{R}^{|V| \times 2}$, vertex indices, $K\in \{1,..,|V|\}^N$:
%, we define a metric $\mathcal{M}$ as follows:
\begin{align}
   \mathcal{M}(\hat{L}, V; K) =\frac{1}{MN}\sum_{m=1}^M\sum_{n=1}^Nz_m||\hat{l}^{2.5D}_{m,n}-{v}^{2.5D}_{m,K_n}||_2
\end{align}
where $z=(h_{box}\times w_{box})^{-\frac{1}{2}}$, the diagonal of the face bounding box. Given the dataset-specific landmark definition, $\tilde{K}$, $\text{NME}=\mathcal{M}(\hat{L}, V; \tilde{K})$.
% 4. We propose our algo to replace those indices to be model specific (a consistency metric). (local consistency alignment metric)
% We replace the dataset-specific landmark definition introduce a normalized mean local consistency metric:
Our normalized mean local consistency metric ($\text{NMLC}$) replaces the dataset-specific landmark definition with a model-specific one:
\begin{equation}
   \text{NMLC}=\min_{K}\mathcal{M}(\hat{L}, V; K),
\end{equation}
enabling fair cross-dataset comparison. Unlike $\text{NME}$, consistent local bias w.r.t. a desired landmark definition, $\tilde{K}$, will not be penalized.
Trivially, $\text{NMLC} \leq \text{NME}$. Non-triviality of $\text{NMLC}$ is ensured by a large test set with pose, identity, and expression variations.
% 5. Analysis of NMLC: refer to figure for qualitative correlation with our metric, e.g. eyes better fitted
Our $\text{NMLC}$ comparisons correlate with qualitative results, see~\cref{fig:result}, as our method's leading performance appears to be reflected. 

\begin{table*}[t]
\centering

\begin{adjustbox}{width=\textwidth,center}
\begin{tabular}{|r|c|c|c|c|c|c|c|c|c|c|c|c|c|}
\hline
\multicolumn{1}{|c|}{} & \multicolumn{6}{c|}{Face Regions} & \multicolumn{3}{c|}{Pose} & \multicolumn{2}{c|}{Expression} & \multicolumn{2}{c|}{Occlusions} \\
\hline
%--------------------------------------------------------------------------------------------------------------------\\
%         |                  Face Regions                  |        Pose            |    Expression  |   Occlusions  \\
%--------------------------------------------------------------------------------------------------------------------\\
Model            &  full  &contours & brows   &  nose  &  eyes   & mouth   & front   & sided    &atypical &   True  & False   &  True   & False   \\
\hline %---------------------------------------------------------------------------------------------------------------
SynergyNet~\cite{wu2021_synergynet}        &   2.81 &    4.17  &   2.87  & 1.96   &   2.19  &   2.39  &   2.59  &   2.91  &   3.27  &   2.52  &   2.99  &   3.65  &   2.68  \\
3DDFA~\cite{zhu2016_face-align}            &   3.45  &   4.79  &   3.51  & 2.76   &   2.78  &   2.99  &   3.21  &   3.48  &   4.33  &   2.98  &   3.73  &   4.62  &   3.26  \\
3DDFA+~\cite{zeng2023_3d-aware}            &   3.21  &   4.50  &   3.26  & 2.60   &   2.56  &   2.75  &   2.98  &   3.26  &   3.95  &   2.86  &   3.42  &   4.06  &   3.07  \\
3DDFAv2~\cite{guo2020_towards-fast}        &  \2{2.45}  & \2{3.76} &   2.37  & 1.77   &   1.92  &   2.01  & \2{2.33} & \2{2.49} &   2.77  & \2{2.22} &   2.59  &   2.81  & \2{2.39} \\
DAD-3DNet\red{$\bigstar$}~\cite{martyniuk2022_dad-3dheads}&\ig{2.08} &   \ig{2.91} &   \ig{2.27} &\ig{1.54}&\ig{1.68} & \ig{1.77}  &\ig{1.87} &  \ig{2.19} & \ig{2.45} & \ig{1.95} & \ig{2.17}  & \ig{2.21} & \ig{2.06} \\
DAD-3DNet+\red{$\bigstar$}~\cite{zeng2023_3d-aware}       &\ig{2.06} &\ig{2.86} &\ig{2.26} &\ig{1.54}&\ig{1.68} &\ig{1.75} &\ig{1.87} &\ig{2.16} &\ig{2.37} &\ig{1.93} &\ig{2.14} &\ig{2.19} &\ig{2.04} \\
FAN3D~\cite{bulat2017_2d-3d}               &   2.46  &   4.24  &   \2{2.34} & \2{1.58}& \2{1.77} & \2{1.82} &   2.35  &   2.52  & \2{2.64} &   2.26  & \2{2.58} & \2{2.77} &   2.41  \\
\hline %---------------------------------------------------------------------------------------------------------------
Ours            &\1{1.91} &\1{3.13} &\1{2.02} &\1{1.28} &\1{1.33} &\1{1.45} &\1{1.76} &\1{2.00} &\1{2.07} &\1{1.77} &\1{1.99} &\1{2.07} &\1{1.88} \\
% \hline %---------------------------------------------------------------------------------------------------------------
\hline %---------------------------------------------------------------------------------------------------------------
Ours (Resnet50)   &   2.23  &  3.71   &  2.23   &  1.59   &  1.51   &  1.71   &  2.05   &  2.36   &  2.38   &  2.05   & 2.35    &  2.68   & 2.16    \\
Ours (Resnet152)   &   2.06  &   3.46  &   2.11  &   1.36  &   1.45  &   1.53  &   1.89  &   2.17  &   2.18  &   1.88  &   2.17  &   2.51  &   1.99  \\
Ours (MF only)  &   2.06  &   3.62  &   2.10  &\2{1.31} &\2{1.40} &\2{1.45} &\2{1.76} &   2.26  &   2.26  &   1.88  &   2.17  &\2{2.17} &   2.04  \\
Ours (MV only)  &   2.33  &   3.46  &   2.31  &   1.70  &   1.83  &   1.95  &   2.14  &   2.43  &   2.56  &   2.13  &   2.44  &   2.85  &   2.24  \\
Ours (100)      &   2.12  &   3.27  &   2.21  &   1.47  &   1.59  &   1.71  &   1.96  &   2.22  &   2.26  &   1.96  &   2.22  &   2.50  &   2.06  \\
Ours (1k)       &\2{2.00} &\2{3.17} &\2{2.09} &   1.40  &   1.47  &   1.57  &   1.85  &\2{2.09} &\2{2.13} &\2{1.86} &\2{2.09} &   2.24  &\2{1.96} \\
% Ours            &\1{1.91} &\1{3.13} &\1{2.02} &\1{1.34} &\1{1.27} &\1{1.45} &\1{1.76} &\1{2.00} &\1{2.07} &\1{1.77} &\1{1.99} &\1{2.07} &\1{1.88} \\
\hline %---------------------------------------------------------------------------------------------------------------

\end{tabular}
\end{adjustbox}
%\caption{Evaluation against SoTA on DAD3D-Heads (part 1/2).. Ablations}
%\label{tab:dadheads_part1}
%\end{table}

\caption{SoTA evaluation (top) and ablations (bottom) on DAD-3DHeads~\cite{martyniuk2022_dad-3dheads}. We report the $\text{NMLC}$ for each model, when averaging across various facial regions and categories. \{Model\}\red{$\bigstar$} denotes the model was trained on the data samples used for our evaluation, and thus not included in our statements relating accuracy.}
\label{tab:eval_dad3d}
\end{table*}

\paragraph{Comparisons}
%We evaluate our method against on both studio-captured 3D photogrammetric ground-truth data from the Multiface dataset, along with \emph{in-the-wild} 3DMM-labeled data from the DAD-3DHeads dataset.  In order to compare our model against previous SoTA with a different landmark definition, we first introduce a novel 3D landmark evaluation scheme.
We evaluate our method on the studio-captured photogrammetric ground-truth Multiface~\cite{wuu2022_multiface} dataset, along with 3DMM-labeled \emph{in-the-wild} images from DAD-3DHeads~\cite{martyniuk2022_dad-3dheads}. As the Multiface dataset is extremely large (65TB), we select 6 sequences, which cover a range of facial expressions, including asymmetric facial deformations. See supplemental document for curation details. The DAD3D-Heads training set offers category labels for pose, expression, occlusion, quality, lighting, gender, and age. So, we chose the above for our evaluations to obtain fine-grained analysis (quality, lighting, gender, and age reported in supplementary). Since our detector outputs 98 landmarks, 
%instead of 68 like the SoTA methods we compare with, we generate the corresponding 68 landmark subset.
we generate the corresponding 68 landmark subset to compare with SoTA methods.
\begin{figure}
    \centering
    \includegraphics[width=1.0\columnwidth]{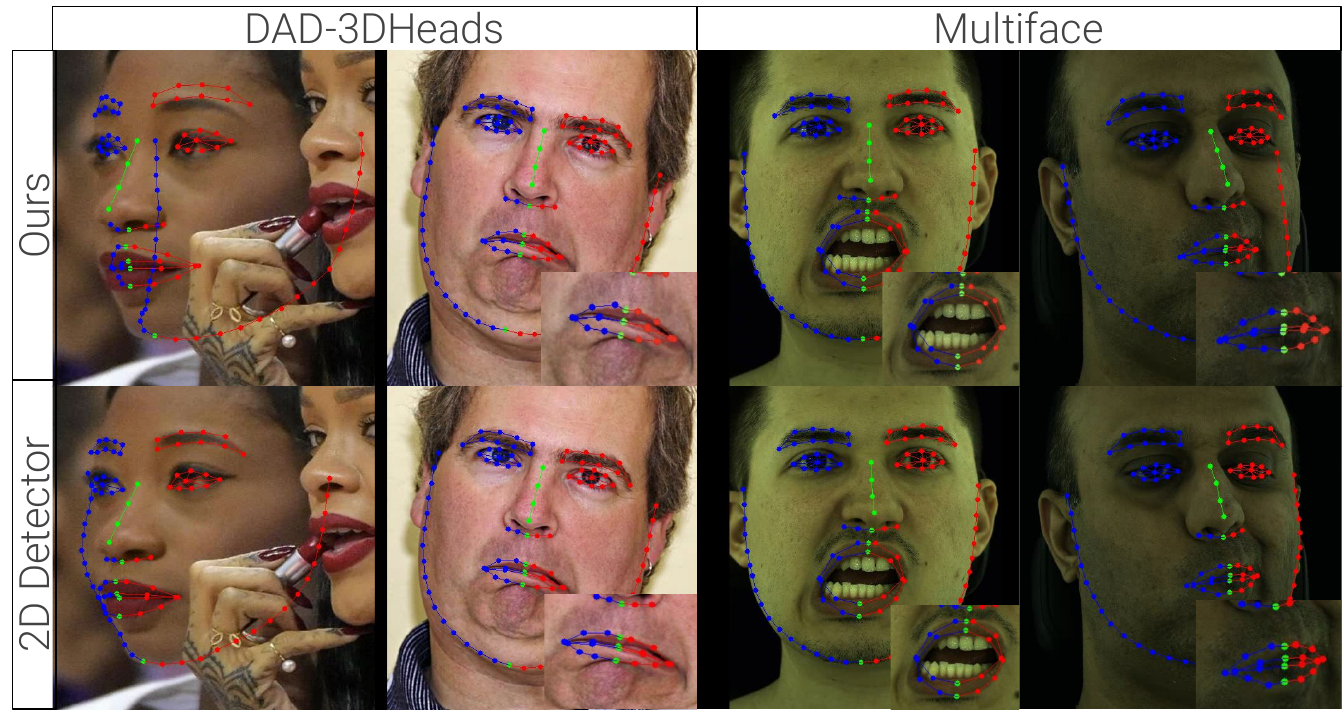}
    \caption{We observe failure cases for our model, including extreme asymmetric expressions and overlapping faces with occlusion, and visualize our model outputs alongside the outputs of the 2D detector used in pseudo-labeling.}
    \label{fig:failurecases}
\end{figure}
We observe that, despite training without ground-truth 3D labels, our method outperforms previous SoTA on each dataset by 22\% and 19\%, as shown in~\cref{tab:eval_dad3d} and~\cref{tab:eval_multiface}, respectively.
% (1.91 v. 2.63) and (2.52 v. 3.27), respectively.
\cref{fig:result} shows that our method captures more fine-scale details in the eye, mouth, and brow regions, and it can properly handle blinks while other methods fail. 
% Overall, each model receives worse scores on the Multiface dataset than that of DAD-3DHeads dataset due to the extreme pose variability of the former. % multi-view dataset. 
We remark, however, that our method still fails for extreme expressions such as ``puckers'' and asymmetric deformations. We hypothesize that improving the 2D detector for such cases will propagate through our pipeline toward improving 3D results, as suggested by observations in~\cref{fig:failurecases}. Note that the model used for 2D pseudo-labels fails similarly for mouth deformations.

\begin{table}[t]
    \small
    \begin{adjustbox}{width=\columnwidth,center}
    \begin{tabular}[t]{|r|c|c|c|c|c|c|}
        \hline
        Model           &  full   &contours  & brows   &  nose   &    eyes &   mouth \\
        \hline 
        SynergyNet~\cite{wu2021_synergynet}       &    3.59 & 4.59 & 3.16 & 3.05 & 2.66 & 3.74 \\
        3DDFA~\cite{zhu2016_face-align}           &    5.06 & 7.31 & 4.29 & 4.36 & 3.70 & 4.66 \\
        3DDFA+~\cite{zeng2023_3d-aware}           &    4.92 & 6.88 & 4.19 & 4.39 & 3.56 & 4.69 \\
        3DDFAv2~\cite{guo2020_towards-fast}       &    3.25 & 4.46 & 2.63 & 2.57 & 2.34 & 3.39 \\
        DAD-3DNet~\cite{martyniuk2022_dad-3dheads}&    3.30 & 4.33 & 3.21 & 2.80 & 2.54 & 3.16 \\
        DAD-3DNet+~\cite{zeng2023_3d-aware}       &    3.28 & \2{4.29} & 3.18 & 2.85 & 2.53 & 3.10 \\
        FAN3D~\cite{bulat2017_2d-3d}              & \2{3.13}& 5.04 & \2{2.58}&\2{2.14} &\2{2.18} &\2{2.80} \\
        \hline
        Ours            & \1{2.52} & \1{3.49}&\1{2.27}&\1{2.11} &\1{1.57} &\1{2.56} \\
        \hline
        Ours (Resnet50)   &  2.87    & 4.13    & 2.61   &  1.79   & 2.13    & 2.86  \\
        Ours (Resnet152)   & \2{2.62} &    3.75 &\2{2.38}&\2{2.14} &\2{1.60} &\2{2.60}   \\
        Ours (MF only)  &    3.51  &    6.21 &   2.94 &   2.44  &   2.24  &   2.75    \\
        Ours (MV only)  &    2.98  &    3.70 &   2.81 &   2.46  &   2.07  &   3.22    \\
        Ours (100)      &    2.84  &    3.82 &   2.49 &   1.84  &   2.37  &   3.0       \\
        Ours (1k)       &    2.73  & \2{3.69}&   2.49 &   2.31  &   1.72  &   2.84      \\
        % \hline
        % Ours            & \1{2.52} & \1{3.49}&\1{2.27}&\1{1.59} &\1{1.95} &\1{2.56}\\
        \hline 
    \end{tabular}
    \end{adjustbox}
    \caption{SoTA evaluation (top) and ablations (bottom) on Multiface~\cite{wuu2022_multiface}. We report the $\text{NMLC}$ for each model, when averaging across various facial regions.}
    \label{tab:eval_multiface}
\end{table}

\paragraph{Ablation Studies}
We conduct ablation studies to observe the effects of the two data sources independently, our choice of encoder backbone, and the impact of sample size. We train our method with multi-view GAN samples and multi-frame video samples independently, referred to as Ours (MV only) and Ours (MF only), respectively, to observe the strengths and weaknesses of each in isolation. We also train our method with a standard Resnet152~\cite{he2016_deep-residual} backbone, pre-trained on ImageNet~\cite{deng2009_imagenet}, of similar parameter count (60M) to our FaRL~\cite{zheng2022_general-rep} backbone (87M), and refer to this model as Ours (Resnet152). Additionally, we include Ours (Resnet50) to compare with the same backbone used by~\cite{martyniuk2022_dad-3dheads, zeng2023_3d-aware}. Finally, we decrease the number of videos/GAN latents from 10k to 1k and 100 samples, referred to as Ours (1K) and Ours (100). \cref{tab:eval_dad3d} and \cref{tab:eval_multiface} show that, when training without multi-view samples (MF only), the model failures for the Multiface dataset are quite pronounced for the contours, as the method struggles to capture large pose variation without the 3D constraints of the multi-view training, which is accentuated for contour landmarks. When training without multi-frame samples (MV only), we observe a sizable performance decrease for both mouth and occlusions, as its training distribution is limited by the FFHQ-trained GAN. Replacing our ViT backbone with a Resnet152 of similar parameter count yielded a slight drop in performance. Interestingly, occlusions yield a more significant drop. We hypothesize that it is a result of the ViT's global reasoning capacity, ability to selectively ignore occluded regions, and the backbone encoder's (FaRL) face image embedding prior. We observe a trend that performance improves with the number of training pseudo-labels we generate.

%% file: sec/5_conclusion.tex
\section{Conclusion}
In this paper, we have introduced a semi-supervised method for geometric prior-free localization with accurate 3D facial landmarks, aligned with 2D human labels, by exploiting multi-view 3D-aware GANs and using 2D landmarks with no ground-truth 3D dataset. We have shown, for the first time, that SoTA 3D landmarks can be learned without 3D labels, paving the way toward improving 3D facial landmarks beyond the limitations of current 3D labeling techniques.
\paragraph{Limitations \& Future Directions}
Despite the promising results demonstrated by our 3D facial landmark localization method, there are still some limitations. We are heavily dependent on the quality of both a 2D landmark detector and a 3D-aware facial GAN. Improvements to either of these dependencies should result in improvements when training with our approach. Currently, we observe limitations in the 2D landmark detector for facial expressions, such as puckers and asymmetric deformations, constraining the performance of 3D uplifting. However, correcting this may be as simple as labeling such examples when training the 2D detector. As has been noted by previous approaches ~\cite{Wu2023LPFFAP}, 3D-aware GANs have limited pose and expression distributions, limiting their downstream application for multi-view consistency. Noting the failure case observed where the method fails due to an occlusion~\cref{fig:failurecases}, future work may include investigating GAN sample augmentation via volumetrically generated occlusions. Finally, as we observe a strong trend in increasing performance improvement with the number of pseudo-labels generated, future work may explore the asymptotic limits of such improvement.
% \begin{wrapfigure}{r}{0.3\textwidth}
%     \includegraphics[width=0.28\textwidth]{figures/focal.pdf}
%   \caption{Scaling the z-translation proportionally to the focal length, we show varying focal lengths for sampling a landmark, image pair from the 3D GAN.}
%   \label{fig:future_work_focal}
% \end{wrapfigure}
% \todo{subfigure with 6 \& 7}
%As shown by recent research for human pose estimation work~\cite{wang2023_zolly}, which introduced a Blender rendered synthetic dataset for handling perspective distortions, focal awareness can yield improve performance on such distorted images. We show that our method may perhaps be adapted for similar purposes, as shown in \cref{fig:future_work_focal}.

% \begin{figure}
%     \centering
%     \includegraphics[width=0.5\textwidth]{figures/focal.pdf}
%      \caption{Scaling the z-translation proportionally to the focal length, we show varying focal lengths for sampling a landmark, image pair from the 3D GAN. We save exploiting this for future work.}
%     \label{fig:future_work_focal}
% \end{figure}

%% file: sec/6_supplementary.tex
\setcounter{page}{1}
%\setcounter{section}{0}
% \maketitle
\maketitlesupplementary

%%%%%%%%%%%%%%%%%%%%%%%%%%%%%%%%%%%%%%%%%%%%%%%%%%%%%%%%%%%%%%%
\section{Implementation Details}

\paragraph{3D Landmark Transformer Architecture}
\begin{figure*}[ht]
    \centering
    \includegraphics[width=1.0\textwidth]{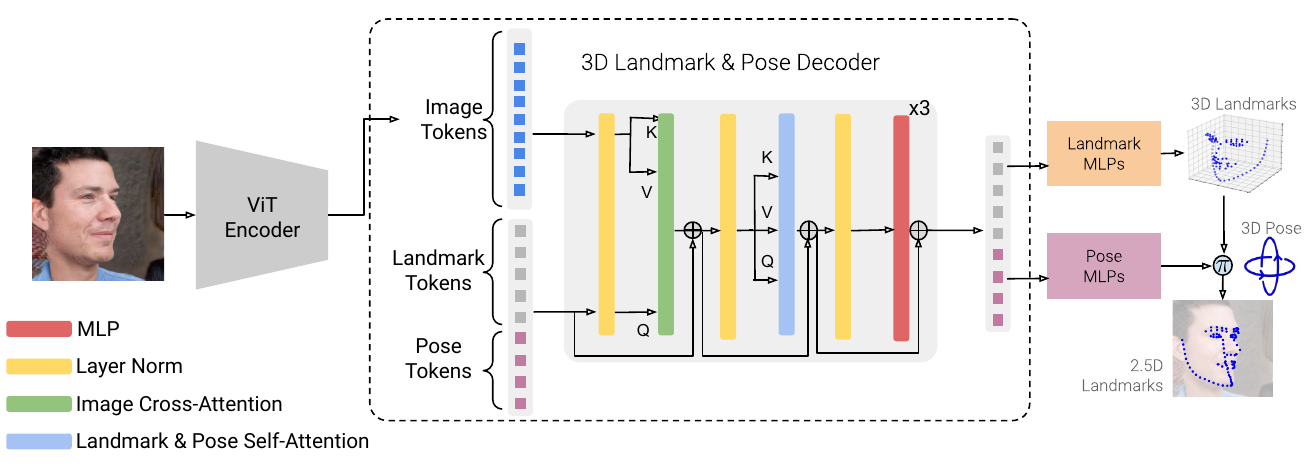}
    \caption{\textbf{3D Landmark Transformer Architecture.}}
    \label{detailed_arch}
\end{figure*}

\cref{detailed_arch} presents a more detailed figure of our 3D landmark transformer architecture. 3D head pose and facial landmarks are estimated via cross-attention and self-attention heads and MLP layers.

\paragraph{Loss Implementation Details}
In order to compute occlusion-aware masks, $m\in \{0,1\}^{N}$, used in Eq. {\color{red} 4}, we apply the predicted rotation matrix to a template of normal vectors for each landmark, and threshold the dot product with the forward vector to obtain the mask. We obtain our normal template by selecting the landmarks on a face mesh, and computing the normals at those locations. We set the threshold so that products above $0.5$ were considered visible, while lowering this threshold to $-0.1$ for the nose bridge. We found this conservative masking strategy reasonable in our experiments.

\paragraph{Multi-view Camera Optimization}
\begin{figure*}[ht]
    \centering
    \includegraphics[width=1.0\textwidth]{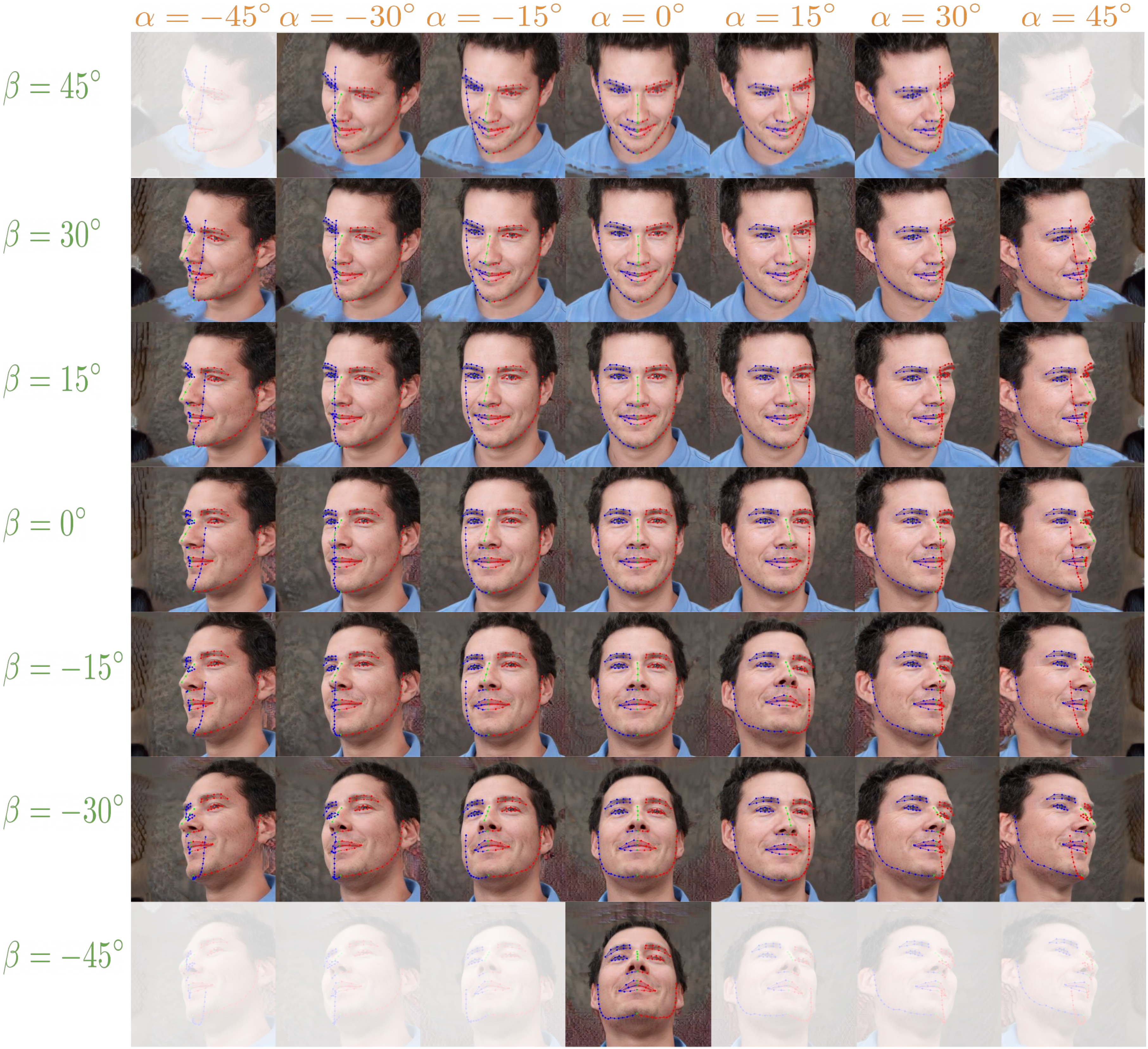}
    \caption{\textbf{Masked Multiview 3D Landmark Optimization's Camera ``Rig'':} Sample camera views used to perform the masked multiview 3D landmark optimization.}
    \label{fig:camera_rig}
\end{figure*}

In obtaining 3D pseudo-labels for 3D-aware GAN-generated samples, we perform a multi-view 3D landmark optimization over detections from renders of camera views, $\tilde{c}_i \in \tilde{C}$, represented by $(\alpha, \beta)$ azimuth and elevation pairs. \cref{fig:camera_rig}~illustrates all these $|\tilde{C}|=41$ sample views.

%%%%%%%%%%%%%%%%%%%%%%%%%%%%%%%%%%%%%%%%%%%%%%%%%%%%%%%%%%%%%%%
\section{Evaluation Set Preparation}
When comparing our model on the DAD3D-Heads~\cite{martyniuk2022_dad-3dheads} dataset, we upsample the meshes to ensure that the mesh is dense enough that the distance between vertices is much smaller than the model's inconsistencies.

Due to the enormous size of the Multiface~\cite{wuu2022_multiface} dataset, we sample a subset for our evaluations. We selected 6 sequences: \emph{Neutral Eyes Open, Relaxed Mouth Open, Open Lips Mouth Stretch Nose Wrinkled, Mouth Nose Left, Mouth Open Jaw Right Show Teeth, Suck Cheeks In}, which include closed eyes, wide mouth openings, and asymmetric facial deformations. The data covers a wide range of cameras, and we discard several in which the face is not visible, including cameras numbered 400055, 400010, 400067, 400025, 400008, and 400070. To eliminate redundancy in the evaluation set, we sample every 15 frames from the downloaded sequences.

%%%%%%%%%%%%%%%%%%%%%%%%%%%%%%%%%%%%%%%%%%%%%%%%%%%%%%%%%%%%%%%
\section{Additional Experiments}

\paragraph{Pseudo-labels Visualized}
\begin{figure*}[ht]
    \centering
    \includegraphics[width=1.0\textwidth]{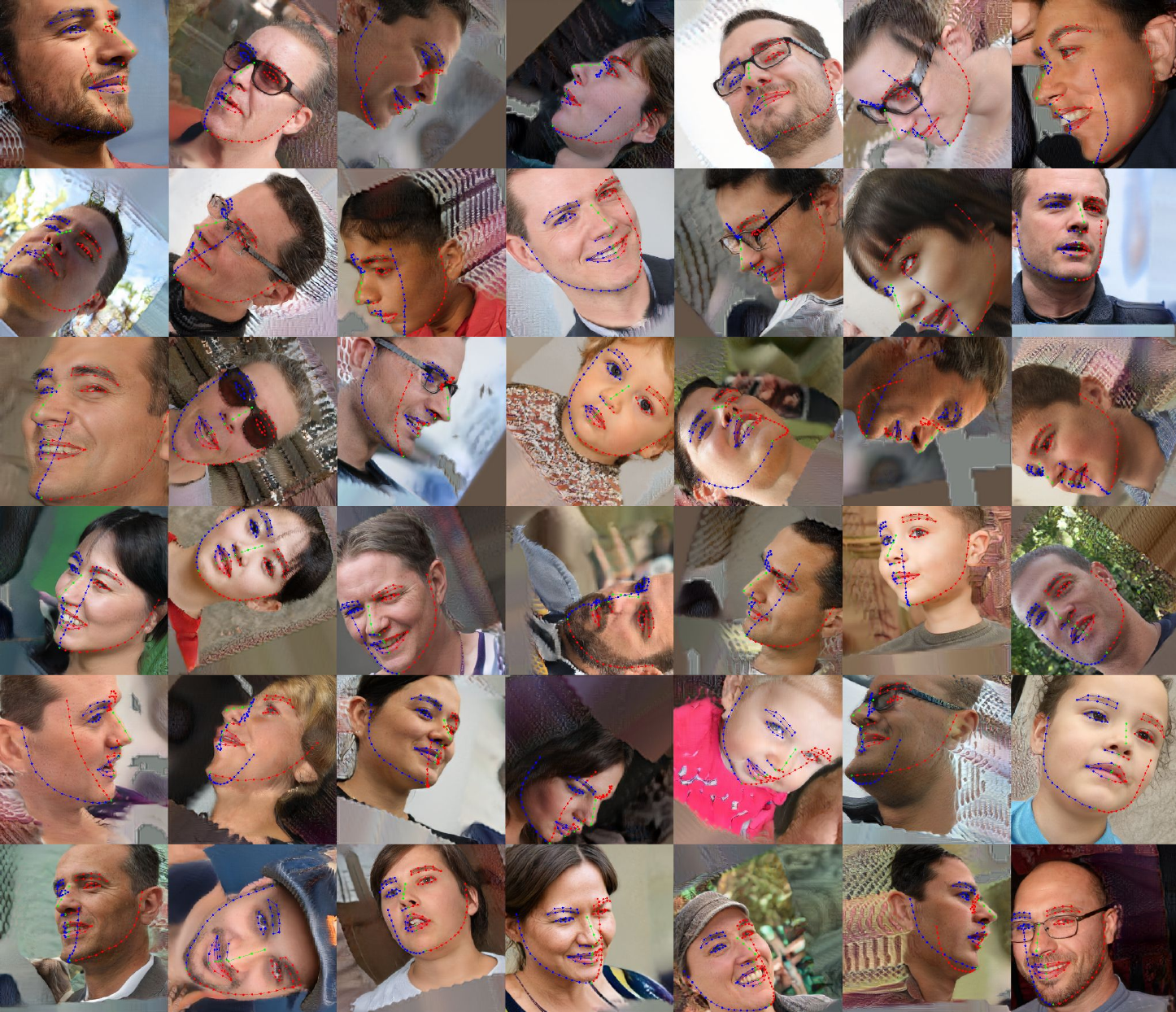}
    \caption{\textbf{3D-aware GAN Pseudo-labeled Samples.} Our approach can faithfully reconstruct 3D landmarks under extreme 3D head poses, and face outline landmarks are not affected by inherent GAN noise around face boundaries.}
    \label{fig:gan_samples}
\end{figure*}
In~\cref{fig:gan_samples}, we visualize 3D-aware GAN samples, obtained via 3D pseudo-labeled IDE-3D~\cite{sun2022_ide} latent renders, which are sampled from our augmented camera space, $\mathbb{C}$.

\paragraph{Additional Qualitative Results on CelebV-HQ}
\begin{figure*}[ht]
    \centering
    \includegraphics[width=1.0\textwidth]{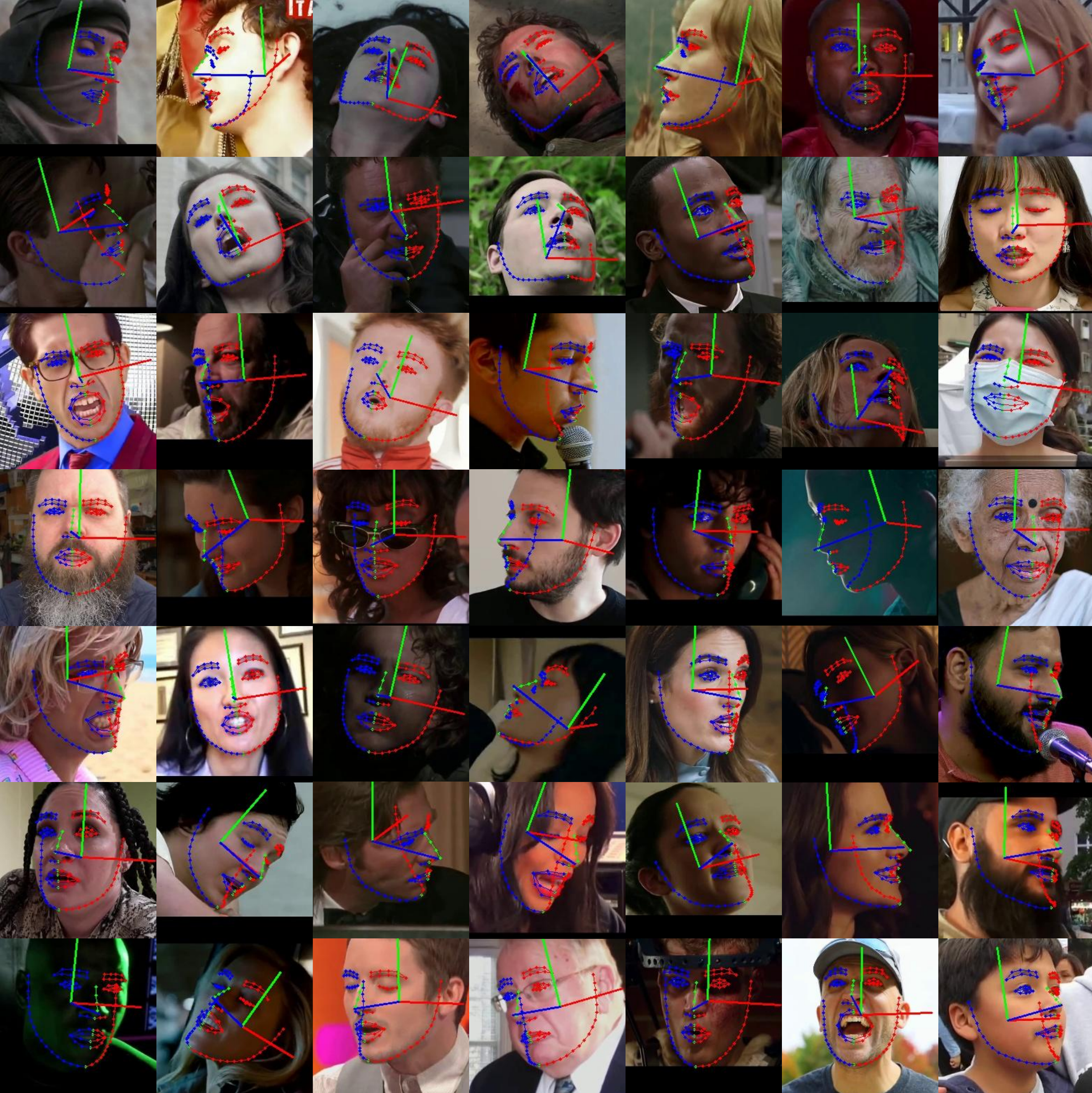}
    \caption{\textbf{Additional Qualitative Results on CelebV-HQ~\cite{zhu2022_celebv} dataset.} Here, the blue, green, and red axes represent Cartesian coordinates and denote the forward, up, and right vectors, respectively. Our approach can faithfully reconstruct 3D landmarks under challenging 3D head poses and harsh lighting.\looseness=-1}
    \label{fig:additional_qualitative}
\end{figure*}

Additional qualitative results on the CelebV-HQ~\cite{zhu2022_celebv} dataset are shown in~\cref{fig:additional_qualitative}.

\paragraph{Evaluations on Additional DAD3D-Heads Categories}
\begin{table*}[ht]
\small
\begin{adjustbox}{width=\width,center}
\begin{tabular}{|c|c|c|c|c|c|c|c|c|c|c|c|c|c|}
\hline
\multicolumn{1}{|c|}{} & \multicolumn{2}{c|}{Quality} & \multicolumn{2}{c|}{Standard Light} & \multicolumn{3}{c|}{Gender} & \multicolumn{4}{c|}{Age} \\
\hline
%---------------------------------------------------------------------------------------------------------
%               |  Quality     |Standard Light |       Gender            |        Age                         |
%---------------------------------------------------------------------------------------------------------
Model           & High     &  Low   &  True   & False   & female  &  male   &undefined& child   & young   &middle aged& senior \\

\hline %--------------------------------------------------------------------------------------------------
SynergyNet~\cite{wu2021_synergynet}&   2.27 & 3.70 & 2.55 & 3.29 & 2.35 & 2.78 & 6.15 & 2.44 & 2.74 & 2.95 & 2.67 \\
3DDFA~\cite{zhu2016_face-align}&       2.80 & 4.50 & 2.95 & 4.34 & 2.99 & 3.41 & 6.71 & 2.94 & 3.45 & 3.53 & 3.30 \\
3DDFA+~\cite{zeng2023_3d-aware}&       2.70 & 4.04 & 2.88 & 3.79 & 2.81 & 3.21 & 5.70 & 2.82 & 3.22 & 3.26 & 3.14 \\
3DDFAv2~\cite{guo2020_towards-fast}&   2.13 & \2{2.97}& 2.33 & \2{2.67}& 2.17 &\2{2.51}&\2{3.77}& 2.23 & 2.45 &\2{2.49}& 2.43 \\
DAD-3DNet\red{$\bigstar$}~\cite{martyniuk2022_dad-3dheads}&\ig{1.84} &\ig{2.48} &\ig{1.99} & \ig{2.26} & \ig{1.87} &\ig{2.15} &\ig{2.88} &\ig{1.83} &\ig{2.05} &   \ig{2.16} &\ig{1.96} \\
DAD-3DNet+\red{$\bigstar$}~\cite{zeng2023_3d-aware}&\ig{1.84} &\ig{2.43} &\ig{1.98} &\ig{2.21} &\ig{1.87} &\ig{2.13} &\ig{2.75} &\ig{1.83} &\ig{2.03} &\ig{2.13} &   \ig{1.97} \\
FAN3D~\cite{bulat2017_2d-3d}&          \2{1.99}& 3.22 &\2{2.21}& 2.91 &\2{2.08}&\2{2.51}& 4.46 &\2{2.02}&\2{2.32}& 2.67 &\2{2.36}\\
\hline %---------------------------------------------------------------------------------------------------------------
Ours            &\1{1.68} &\1{2.28} &\1{1.81} &\1{2.07} &\1{1.72} &\1{1.95} &\1{2.72} &\1{1.70} &\1{1.90} &\1{1.95} &\1{1.85} \\
\hline %---------------------------------------------------------------------------------------------------------------
Ours (Resnet50)    &   1.92  &   2.75  &   2.11  &   2.45  &   2.03  &   2.23  &   3.55  &   1.91  &   2.24  &   2.29  &   2.07  \\
Ours (Resnet152)   &   1.81  &   2.47  &   1.96  &   2.23  &   1.87  &   2.08  &   3.05  &\2{1.80} &   2.07  &   2.09  &   1.98  \\
Ours (MF only)  &   1.80  &   2.48  &   1.96  &   2.24  &   1.87  &   2.12  &\2{2.78} &   1.83  &   2.10  &   2.08  &\2{1.88} \\
Ours (MV only)  &   2.01  &   2.85  &   2.17  &   2.60  &   2.10  &   2.33  &   3.75  &   2.03  &   2.36  &   2.35  &   2.21  \\
Ours (100)      &   1.89  &   2.50  &   2.02  &   2.30  &   1.93  &   2.14  &   3.13  &   1.92  &   2.14  &   2.14  &   2.03  \\
Ours (1k)       &\2{1.78} &\2{2.37} &\2{1.91} &\2{2.17} &\2{1.81} &\2{2.04} &   2.89  &   1.82  &\2{2.01} &\2{2.03} &   1.94  \\

\hline %-----------------------------------------------------------------------------------
\end{tabular}
\end{adjustbox}
\caption{SoTA evaluation (top) and ablations (bottom) on DAD-3DHeads~\cite{martyniuk2022_dad-3dheads}, for additional categories. We report the $\text{NMLC}$ for each model when averaging across various facial regions and categories. \{Model\}\red{$\bigstar$} denotes the model was trained on the data samples used for our evaluation.}
\label{tab:eval_dad3d_additional}
\end{table*}

In~\cref{tab:eval_dad3d_additional}, we report the DAD3D-Heads~\cite{martyniuk2022_dad-3dheads} evaluation results for additional categories, including image quality, lighting, gender, and age.

\paragraph{Loss Function Ablations}
We compare the loss function used by our method, Laplacian Log Likelihood, with other common loss functions, L1 and MSE, in~\cref{tab:eval_loss_ablation}. Our choice yields the best results on our benchmark datasets.

\begin{table}[t]
    \small
    \begin{adjustbox}{width=\columnwidth,center}
    \begin{tabular}[t]{|r|c|c|c|c|c|c|}
        \hline
        Model           &  Multiface   & DAD3D-Heads  \\
        \hline 
        Ours (LLL Loss)      &  \1{2.52}    & \1{1.68}    \\
        Ours (L1 Loss)       &  2.82    & 2.08    \\
        Ours (MSE Loss)      &  3.01    & 2.49    \\
        \hline 
    \end{tabular}
    \end{adjustbox}
    \caption{Ablation studies concerning the loss function used, where Ours uses Laplacian Log Likelihood (LLL), evaluated on the full set of landmarks from Multiface~\cite{wuu2022_multiface} and DAD3D-Heads~\cite{martyniuk2022_dad-3dheads}. We report the $\text{NMLC}$ for each model, when averaging across various facial regions.}
    \label{tab:eval_loss_ablation}
\end{table}

\paragraph{Cross-Dataset Evaluations}
Our investigations into cross-dataset evaluations reveal a notable limitation in model generalizability between datasets with differing labeling conventions. We report cross-dataset evaluations in~\cref{tab:cross_dataset_eval} on both AFLW2000-3D~\cite{zhu2016_face-align} and the DAD3D-Heads~\cite{martyniuk2022_dad-3dheads} validation set, comparing our method with methods trained on DAD3D-Heads and 300WLP~\cite{zhu2016_face-align}, noting that 300WLP's compatible evaluation set is the AFLW2000-3D dataset. We observe that despite a global alignment in how landmarks are defined, cross-dataset scores of every SoTA model are all worse than the SoTA models of the compatible dataset. This is expected due to the local definition bias w.r.t. a different dataset's landmark definition, which yields a consistent error. For each dataset, our model achieves the best cross-dataset score. The cross-dataset metrics do not disentangle the local definition bias from some notion of actual error with respect to the model's landmark definition. Intuitively, if our model's landmark definition were the midway interpolation between the two dataset definitions, our model would incur half of the error from local definition bias than that of the other models. Hence, for fair comparisons, we compare against other methods using our proposed $\text{NMLC}$ metric, which removes the local definition bias from the evaluated error. Nevertheless, cross-dataset evaluation remains a useful proxy for assessing the global consistency of landmark definitions across models, a presupposition integral to the $\text{NMLC}$ metric.

\begin{table*}[ht]
\small
\begin{adjustbox}{width=\textwidth,center}
\begin{tabular}{|c|c|c|c|}
\hline
\multicolumn{1}{|c|}{Model} & \multicolumn{1}{|c|}{Training Set}  & \multicolumn{1}{c|}{AFLW2000-3D-reannotated NME} & \multicolumn{1}{c|}{DAD3D-Heads NME} \\
\hline
%---------------------------------------------------------------------------------------------------------
\textbf{}%    Model   | Training Set                     | AFLW2000-3D-reannotated NME  |    DAD3D-Heads     |
%---------------------------------------------------------------------------------------------------------
% 3DDFA~\cite{zhu2016_face-align}            & 300WLP        & 4.30 & 3.91\checkmark  \\
% 3DDFA+~\cite{zeng2023_3d-aware}            & 300WLP        & 4.26 & 3.77\checkmark \\
FAN3D~\cite{bulat2017_2d-3d}               & 300WLP        & 2.85 & 3.83\checkmark \\
SynergyNet~\cite{wu2021_synergynet}        & 300WLP        & 2.65 & 3.46\checkmark  \\
3DDFAv2~\cite{guo2020_towards-fast}        & 300WLP        & 3.33 & 3.10\checkmark  \\
DAD-3DNet~\cite{martyniuk2022_dad-3dheads} & DAD3D-Heads   & 5.10\checkmark & 2.71 \\
DAD-3DNet+~\cite{zeng2023_3d-aware}        & DAD3D-Heads+  & 5.00\checkmark & 2.71  \\
\hline
Ours                                       & FaceLift      & 3.51\checkmark & 2.78\checkmark  \\

\hline
\end{tabular}
\end{adjustbox}
\caption{Cross-dataset evaluation of $\text{NME}$ on AFLW2000-3D-reannotated~\cite{zhu2016_face-align} and the validation set of DAD3D-Heads~\cite{martyniuk2022_dad-3dheads}. \checkmark denotes that the score is cross-dataset, meaning the training set definition is not compatible with the evaluation dataset and definition. We see that while our model is the best on the cross-dataset comparisons for each dataset, compatible SoTA models yield better scores since they do not incur the local definition bias of cross-dataset evaluation.}
\label{tab:cross_dataset_eval}
\end{table*}